\documentclass[journal,twoside]{IEEEtran}
\usepackage[cmex10]{amsmath}
\usepackage{amssymb}
\usepackage{amsfonts}
\usepackage{amsthm}
\usepackage{algorithm}
\usepackage{algpseudocode}
\usepackage{graphicx}
\usepackage{epsf}
\usepackage{physics}
\usepackage[utf8]{inputenc}

\graphicspath{{figures/}}
\usepackage{ wasysym }

\usepackage{epsfig}
\usepackage{cite}
\usepackage{tensor}
\usepackage[caption=false,font=footnotesize]{subfig}
\usepackage{color,soulutf8}
\usepackage{xcolor}

\usepackage[pdfa]{hyperref}
\hypersetup{
    colorlinks=true,
    linkcolor=blue,
    filecolor=magenta,      
    urlcolor=cyan,
}
\graphicspath{{figures/}}
\theoremstyle{definition}

\renewcommand{\baselinestretch}{0.99}

\begin{document}

\title{Robotics Meets Cosmetic Dermatology: Development of a Novel Vision-Guided System\\for Skin Photo-Rejuvenation}

\author{Muhammad Muddassir,~\IEEEmembership{Student~Member,~IEEE}, Domingo Gomez, Luyin Hu,\\ Shujian Chen and David Navarro-Alarcon,~\IEEEmembership{Senior~Member,~IEEE}%
	\thanks{This work is supported in part by the Research Grants Council of Hong Kong under grant PolyU 142039/17E and in part by RODS Tech Ltd.}%
	\thanks{The authors are with The Hong Kong Polytechnic University, Department of Mechanical Engineering. Corresponding author e-mail: {\texttt{\small dna@ieee.org}}}%
}

\bstctlcite{IEEEexample:BSTcontrol}

\markboth{IEEE/ASME Transactions on Mechatronics}%
{Muddassir \MakeLowercase{\textit{et al.}}:A Novel Robotic System for Skin Photo-Rejuvenation}
\maketitle

\begin{abstract}
	In this paper, we present a novel robotic system for skin photo-rejuvenation procedures, which can uniformly deliver the laser's energy over the skin of the face. 
	The robotised procedure is performed by a manipulator whose end-effector is instrumented with a depth sensor, a thermal camera, and a cosmetic laser generator. 
	To plan the heat stimulating trajectories for the laser, the system computes the surface model of the face and segments it into seven regions that are automatically filled with laser shots. 
	We report experimental results with human subjects to validate the performance of the system. 
	To the best of the author's knowledge, this is the first time that facial skin rejuvenation has been automated by robot manipulators.
\end{abstract}

\IEEEpeerreviewmaketitle

\section{INTRODUCTION}\label{sec:intro}
\IEEEPARstart{T}here are two main ageing processes that affect a person's skin condition \cite{Journals:Holck2003}.
Ageing due to the biological clock, which up to this day is considered to be irreversible, and photo-ageing, which results from exposure to ultraviolet radiation coming from the sun; The latter is widely considered to be treatable, to some extent \cite{Oblong2009, NARURKAR2009281, OBLONG2009301}. 
With the aim of ``reversing'' skin damage, in the past decades, people have turned to the so-called \emph{beauty clinics} for receiving various types of non-invasive dermatological procedures.
These treatments are typically performed with cosmetic instruments based on laser light \cite{Goldberg1997}, intense pulsed light \cite{Journals:Babilas2010}, radio-frequency \cite{Journals:Lolis2012}, etc.
Worldwide, the beauty industry has seen an exponential increase in the demand for aesthetic skin rejuvenation treatments.

\begin{figure}
	\centering
	\includegraphics[width=\linewidth]{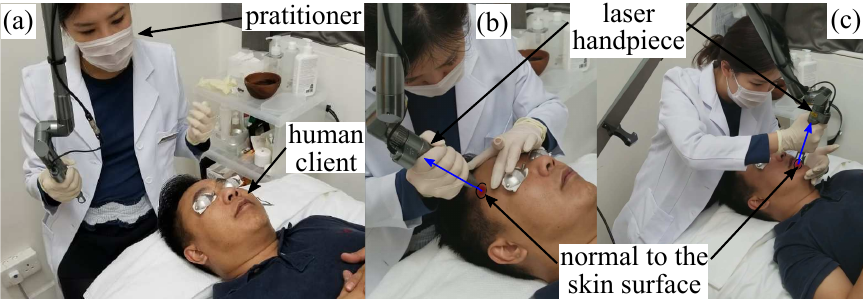}
	\caption{Conventional skin photo-rejuvenation procedure: (a) examination of the facial skin condition, (b) manipulation of the handpiece over the forehead, (c) manipulation of the handpiece on the left jaw.}
	\label{fig:trad_proc}
\end{figure}

A typical rejuvenation procedure conducted in these beauty clinics is shown in Fig. \ref{fig:trad_proc}, where the dermatologist (or ``skin technician'' \cite{Journals:RESNECK2008}) visually examines the skin condition to determine the type of treatment to be performed along with the appropriate laser light parameters \cite{Journals:Eldomyati2011}.
A non-ablative instrument is then manipulated with repetitive motion patterns over different areas of the face to stimulate the skin tissues.
It must be activated with the \emph{exact amount} of energy and time to produce the expected result without causing damage \cite{Goldberg1999}. 
The complete rejuvenation procedure lasts around 25 minutes---It is a tedious and tiring task for practitioners, who must perform it several times in a single day, a situation that contributes to the existing high turnover rate of experienced professionals in the industry \cite{RESNECK200450}.
These issues clearly show the need to develop robotic systems that can automate the manipulation of instruments.

Cosmetic dermatology is currently an under-explored application field in robotics as compared to medical robots \cite{nathoo2005touch,berkelman2004body, haidegger2008future,kwoh1988robot,beasley2012medical, MHwang2019, CLi2020, Vazquez2019}. 
Thus, it has the potential to presents more interesting challenges and opportunities.
Note that there are indubitably few commercially available robotic systems for cosmetic and dermatological treatment \cite{Journals:Draelos2011}.
One of such commercialised systems is ARTAS Follicular Unit Extraction (FUE) robot that can remove healthy follicles from a donor and autonomously transplant them onto the recipient's scalp \cite{bernstein2012integrating}.
Another example is reported in \cite{Lim2014}, who reported a dermatology system to perform hair removal; This system can automatically conduct the laser hair removal by using a manipulator while activating a laser instrument over regions defined by a supervising practitioner/dermatologist \cite{Lim2014, Lim2015, Lim2017, park2015method, Koh2017}.
However, none of these robotic systems is specifically designed for facial skin rejuvenation procedures.

To provide a feasible solution to the open problem, in this article we present an innovative system capable of autonomously performing the skin rejuvenation treatment on a human faces.
The developed system is composed of a 6-DOF robot manipulator with the custom-built end-effector that carries a laser cosmetic instrument.
This system is equipped with a RGB-D camera to reconstruct the facial geometric model and three distance sensors to monitor the surroundings of the end-effector.
To the best of the author's knowledge, this is the first time that a robotic approach for facial skin rejuvenation has been reported in the literature.
The original contributions are: 
\begin{enumerate}
	\item Development of a specialised mechanical prototype for cosmetic procedures.
	\item Design of a new sensor-based method for controlling the stimulation of skin tissues.
	\item Experimental validation of the developed robotic system.
	\item Perform the photo-rejuvenation treatment with proposed robotics system on the human subjects.
\end{enumerate}

The rest of the manuscript is organised as follows: Sec. \ref{sec:pros_rob_sys} presents the proposed prototype; Sec. \ref{sec:sens_sys} describes the sensing system; Sec. \ref{sec:contr_sys} introduces the control algorithm; Sec. \ref{sec:results} reports the experiments; Sec. \ref{sec:conclusions} discusses the conclusions and the future work.

\begin{figure}
	\centering
	\includegraphics[width=\columnwidth]{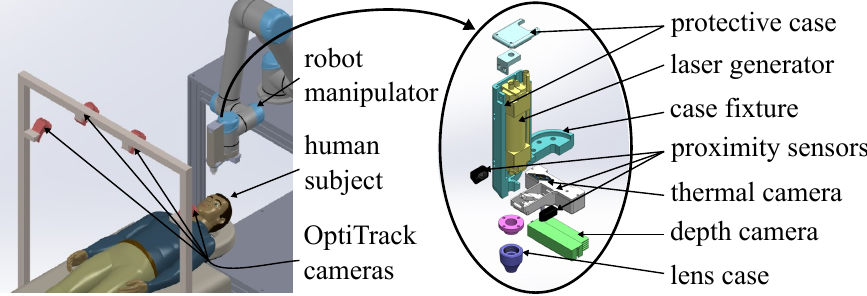}
	\caption{(on left) Proposed setup. Robotised facial skin rejuvenation system. (on right) Exploded view of the customised end-effector}
	\label{fig:setup}
\end{figure}

\section{Robotising Skin Photo-Rejuvenation}	\label{sec:pros_rob_sys}

\subsection{Common Practice}
The operational principle of skin photo-rejuvenation procedure is the thermal stimulation of the collagen in the skin \cite{OBLONG2009301}.
It is done by transferring the energy of the laser light into the skin tissue, which is a slow process of increasing temperature due of accumulation of the laser energy in the skin tissue.
Fig. \ref{fig:trad_proc}(a) depicts the conventional setup to preform skin photo-rejuvenation at the beauty clinics.
The procedure starts with the close examination of the client's skin by a practitioner, then a practitioner empirically sets the parameters of the laser generator machine based on the skin tone and current condition.  
These parameters include laser diameter, laser energy and fluence (laser energy per centimetre square $J/cm^{2}$)\cite{Farkas2013}. 
Fig. \ref{fig:trad_proc}(b) and \ref{fig:trad_proc}(c) show how the laser handpiece is manipulate in a normal direction to the skin surface.
To operate on a particular skin region, the practitioner typically follows an S-shaped path and uses a foot pedal to switch the laser on/off.
Throughout this manuscript, we refer to the instance of delivering the laser light energy on the skin surface as a ``laser shots''.

While delivering the laser light energy, the practitioner should monitor the traces of each laser shots to avoid the overlapping and gaps between each laser shot.
If the laser shots could not distribute uniformly, then the overlapping of laser shots can cause serious side effects (i.e. erythema, hyper-pigmentation, and crusts), whereas the gaps between each laser shot can lead to a sub-optimal stimulation of the skin.
In our study, we have used a Q-switched Nd: YAG (1064 nm) laser, with a pulse duration of <10 ns, and an adjustable repetition rate of 1--10 Hz.
The wavelength of 1064 nm lies within the infra-red spectrum (invisible to the human eye).
The laser generator machine used in this study fires a low energy flash of visible light with each laser shot for the convenience of an operator.
Even with this visual aids, an experienced practitioner can easily lose the track of the degree of the stimulation induced onto the skin.

\subsection{Proposed Setup}
Fig. \ref{fig:setup} conceptually illustrates the proposed robotic system and its various components.
A robot manipulator is placed on top of a wheeled platform that provides the mobility to the system, which is an essential feature for the environments like beauty clinic.
A custom-made rejuvenation end-effector is also developed for the robotic system.
The purpose of this end-effector is to carry all the sensors and laser handpiece during the procedure.
This end-effector was designed considering the following requirements:
\begin{itemize}
 	\item The moving end-effector should have a proximity like behaviour to avoid any possible collision with the human face during the treatment. 
	\item The structure should be rigid enough to hold the laser generator filament, vision and distance sensors during manipulation.
	\item The laser should be able to stimulate any point over the surface of the face.
	\item The structure should be compact.
\end{itemize}
To fulfil the above requirements, a 3D printed rectangle shell structure is designed to house the laser generator and the sensors, as shown in Fig. \ref{fig:setup}.
An extension beside the shell, which have the screw interfaces to physically attached the instrumented customised end-effector to the robot manipulator.
The small extended structures around the shell are the TOF 10120 distance sensors to continuously monitor the closeness of the end-effector to the obstacle in the environment, as in Fig. \ref{fig:setup}. 
The proposed prototype utilises an Orbbec Astra Mini S depth camera for facial model reconstruction task. 
For the sake of simplicity, we called the custom-made end-effector for the photo-rejuvenation treatment as an end-effector of a robot manipulator throughout this manuscript.
To manipulate the laser generator over the facial skin cosmetic instrument, a UR5 manipulator from Universal Robots is used.
The control box of the robot is placed inside the mobile platform; The Linux PC communicates with the robot's servo controller via a TCP/IP socket.
All the proposed algorithms run in the Linux PC with ROS \cite{quigley2009ros}.
The complete system can be divided into two main subsystems: a sensing system and a control system.
Both systems are interconnected with each other and are responsible for dedicated tasks.
A user interface is also developed so that the practitioner can operate the system with a control screen.
In this article, we have considered the coordinate frame of the robot base as an inertial frame of the developed system.

\section{SENSING SYSTEM}	\label{sec:sens_sys}

\subsection{Facial Geometric Model Computation}
The facial surface model reconstruction starts with a search of a human face in an RGB image from the scene. 
A face detector by \cite{king2012dlib} provides a bounding box around a human face.
Then the system enforces the facial landmark detector by \cite{kazemi2014one} to search the landmarks only inside the bounding box to locate the position of the key facial landmarks (i.e. eyes, eyebrows, nose, lips and face boundary) from the captured RGB image.
To compute the pose of the detected face, the system requires the position and the orientation of the face in the camera coordinate frame.
The position of the detected face is assumed to be equivalent to ${}^{C}\vec{r}_{F}$ of a $3 \time 1$ vector.
Lets, $\vec{d}_{l}$ and $\vec{d}_{r}$ are the position vectors of $3 \times 1$ defining the position of the left and right eye of the detected face in a camera coordinate.
Then the position vector of the detected face will be  ${}^{C}\vec{r}_{F} = (\vec{d}_{l}$ + $\vec{d}_{r}) / 2$.
As, the rotation matrix of a coordinate frames consist of three orthonormal column vectors of $3 \times 1$, i.e. $R = [\vec{\alpha}\ \vec{\beta}\ \vec{\gamma}] \in SO(3)$.
The rotation matrix ${}^{C}R_{F}$ that defines the orientation of the face in the camera coordinate frame can be computed by obtaining three orthonormal vectors as follows:
\begin{equation}
{}^{C}R_{F} = 
\begin{bmatrix}
\vec{\alpha}_{F} & \vec{\beta}_{F} & \vec{\gamma}_{F}
\end{bmatrix}
\end{equation}
where $\vec{\alpha}_{F} = \frac{\vec{d}_{r} - {}^{C}\vec{r}_{F}}{\| \vec{d}_{r} - {}^{C}\vec{r}_{F}\|}$, $\vec{\beta}_{F} = \frac{[0\ 0\ 1]^{T} \times \vec{\alpha}_{F}}{\|[0\ 0\ 1]^{T} \times \vec{\alpha}_{F}\|}$ and $\vec{\gamma}_{F} = \frac{\vec{\alpha}_{F} \times \vec{\beta}_{F}}{\|\vec{\alpha}_{F} \times \vec{\beta}_{F}\|}$.
The reason to equate the $\vec{\beta}_{F}$ with the cross product of the  $\vec{\alpha}_{F}$ and $[0\ 0\ 1]^{T}$ is to constrained the $\vec{\beta}_{F}$.
Otherwise, $\vec{\beta}_{F}$ can be defined by the infinite many normal vectors on the plane perpendicular to the $\vec{\alpha}_{F}$. 
Whereas, $[0\ 0\ 1]^{T}$ depicts the last column of the rotation matrix or $\vec{\gamma}$ of the observing coordinate frame.
Now the pose of the face in the camera coordinate is:
\begin{equation}
{}^{C}T_{F} = 
\begin{bmatrix}
{}^{C}R_{F} &    {}^{C}\vec{r}_{F}\\
0           &    1 \\
\end{bmatrix}
\end{equation}
where the ${}^{C}R_{F}$ is a $3 \times 3$ rotation matrix and ${}^{C}\vec{r}_{F}$ is a $3 \times 1$ translation vector. 
After estimating the pose of the face in the camera coordinate, it is transformed to the robot's base coordinate using:
\begin{equation}
{}^{B}T_{F} = {}^{B}T_{E} {}^{E}T_{C} {}^{C}T_{F}
\end{equation}
Here ${}^{B}T_{F}$ defines the pose of the face in robot's base coordinate, ${}^{B}T_{E}$ is the homogeneous transformation from robot's base to end-effector. 
${}^{C}T_{F}$ is defining the pose of the face in camera coordinate.

\begin{figure}
	\centering
	\includegraphics[width=\linewidth]{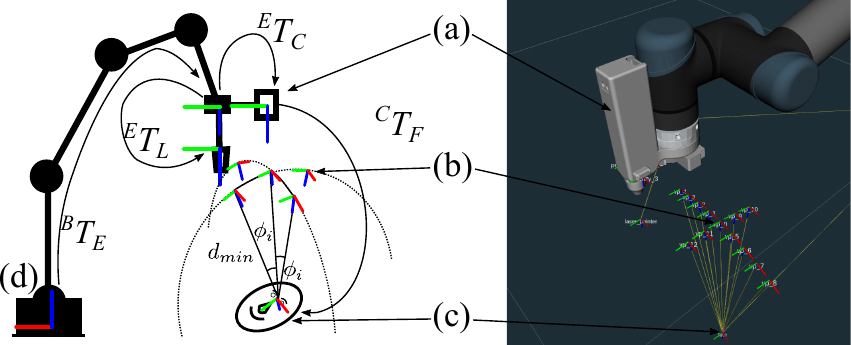}
	\caption{Estimating the viewpoints. (a) depth camera. (b) estimated viewpoints ${}^{B}T_{v_{i}}$. (c) pose of the human face ${}^{B} T_{F}$. (d) coordinate frame of the robot base (inertial frame of reference)}
	\label{fig:viewpoints}
\end{figure}

The commercial depth cameras have a limited field of view (FOV), thus the depth data acquired from one view may contain inconsistencies (or holes).
To reconstruct the consistent (smooth) facial model, the sensing system estimates $N_{v}$ number of viewpoints around the detected face in order to acquire the depth data from different viewpoints (or vantage points).
The capturing of the images from different viewpoints avoids the possible occlusion and provides detailed surface of a human face.
These viewpoints are estimated with the increments and decrements of a predefined angle $\phi_{i}$ around the pose of the detected face ${}^{B}T_{F}$, the angle $\phi_{i}$ will be same for both latitudinal (around $x$-axis of the ${}^{B}T_{F}$) and longitudinal viewpoints(around $y$-axis of the ${}^{B}T_{F}$), as shown in Fig \ref{fig:viewpoints}. 
It is assumed that the longitudinal viewpoints will lie on the circles at $y=0$ in the $zx$-plane.
Thus can be estimated using the parametric equation of circle, as ${}^{F}R_{v_{i}} = R_y(\phi_i) $ and ${}^{F}\vec{r}_{v_{i}} = [-d_{min} \sin(\phi_i), 0, -d_{min}]$, resulting a homogeneous transformation matrix ${}^{F}T_{v_{i}}$ of $4\times4$.
In inertial frame (robot base) it will be ${}^{B}T_{v_{i}} = {}^{B}T_{F} {}^{F}T_{v_{i}}$.
Where, the $d_{min}$ in the Fig. \ref{fig:viewpoints} defines the minimum range of the depth camera.
Similarly, the latitudinal viewpoints is estimated around $x$-axis on the $yz$-plane, by ${}^{F}R_{v_{i}} = R_x(\phi_i)$ and ${}^{F}\vec{r}_{v_{i}} = [0, d_{min} \sin(\phi_i), -d_{min} \cos(\phi_i)]^T$ and in inertial frame it will be ${}^{B}T_{v_{i}} = {}^{B}T_{F} {}^{F}T_{v_{i}}$.
The estimated viewpoints around the face are shown in Fig. \ref{fig:viewpoints}.


To acquire the visual data from the estimated viewpoints, the control system commands the end effector to visit each estimated viewpoint ${}^{B}T_{v_{i}}$.
After reaching a viewpoint, sensing system captures the visual data and extracts only the region having the detected face from the captured RGB and depth images.
Then, these regions from both images are converted into the point cloud.
Each point cloud has an offset transformation in their origin to the point clouds captured from other viewpoints.
To align all the point clouds, the relative pose among each viewpoint should be known.
As the control system receives continuous feedback of end-effector position from the robot manipulator. 
The relative pose between each viewpoint can be computed as ${}^{v_{i}}T_{v_{i+1}} = ({}^{B}T_{v_{i}})^{-1} {}^{B}T_{v_{i+1}}$.
Additionally, the Point to Plane ICP \cite{low2004linear} followed by CICP \cite{park2017colored} is applied to the point clouds captured from each viewpoint.  
Furthermore, to remove the noisy data points, the voxel grid down-sampling, with voxel leaf length of $2mm$, is performed on the final facial model.
This down-sampling keeps the density consistent and remove the holes from the final facial model.
Each point in a point cloud is a structure of three vectors, denotes as $\{\vec{x}, \vec{n}, \vec{c}\}$, where $\vec{x} = [x\ y\ z]^{T}$ is a position vector, $\vec{n} = [n_{x}\ n_{y}\ n_{z}]^{T}$ represents the normal vector and $\vec{c} = [r\ g\ b]^{T}$ defines the colour.

\subsection{Automatic Face Segmentation}
\begin{figure}
	\centering
	\includegraphics[width=\columnwidth]{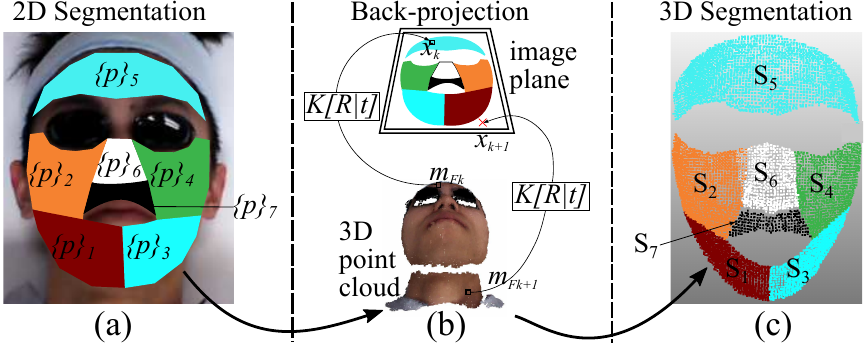}
	\caption{Segmentation of facial model using 2D facial landmarks. (a) polygons $\{p\}_{j}$ defines each region. (b) back-project the 3D point to 2D image plane. (c) segments $M_{F}$ point cloud into $S_{j}$ segments, where $j=1,2,\dots 7$}    \label{fig:back_proj}
\end{figure}

The adipose layer beneath the facial skin is uneven throughout the face, e.g. the cheek region has thicker fats layer than the forehead region.
Therefore, defining each region into segments can aid to set the appropriate laser light energy level, depending on the region to be stimulated.
Furthermore, the classification of the different parts of a face enables to compute trajectory separately for the laser instrument that smoothly follows the curved surface of the skin with the required normal orientation to it; Note the if such geometric constraints are not considered within the trajectory generation, the manipulated instrument can potentially collide with protruding areas of the face (e.g. the nose). 
To cope these issues, an algorithm to automatically segment the human face into seven regions is also proposed, as in Fig. \ref{fig:back_proj}.
The classification of each region of a face is called ``segmented region'' in this manuscript, which are comprised of the left cheek, left jaw, right cheek, right jaw, forehead, nose and upper lips.


The initial facial model that is computed with the depth camera is in the form of a point cloud of unorganised 3D data (where neighbouring points in space are not necessarily adjacent in computer memory indexing, and vice-versa).
To plan a path for a robot manipulator which can perform the photo-rejuvenation over the surface, the control system should aware of the correspondence of each point in a model to each segment of a face. 
As a solution to this problem, we developed an algorithm that uses 2D facial landmarks to cluster the unstructured data into seven segments (point clouds), see Fig. \ref{fig:back_proj}.
Lets the $S_{j}$ (for $j=1,\dots,7$) denotes the $j$th region of facial model and an arbitrary $i$th point in $S_{j}$ is $s_{i}^{j}$.

Initially, the auto-segmentation algorithm detects the facial landmarks using \cite{kazemi2014one, king2012dlib}, from RGB image captured from the first viewpoint ${}^{B}T_{v_{0}}$, which provides the most reliable view to detect the facial landmarks (hence no facial feature can occlude from the front view).
Once these key landmarks have been detected, the algorithm draws polygons $\{p\}_{j}$ on the image plane to define the seven regions on the image plane, as shown in Fig. \ref{fig:back_proj}(a).
Then, a sorting routine runs seven parallel processes to back-project each 3D point $m_{F_{k}}$ of the facial model $M_{F}$ on the image plane, by using the camera's perspective projection relation:
\begin{equation}
\begin{bmatrix} \vec{\xi} \\ 1 \end{bmatrix} = K[R | t] \begin{bmatrix} \vec{x} \\ 1 \end{bmatrix}
\end{equation}
Here $\vec{\xi}$ is a $2 \times 1$ column vector and denotes the projection of a 3D point $\vec{x}$ on an image plane.
Where $K$ as the $3 \times 3$ matrix of intrinsic parameters, and $R,t$ as the extrinsic parameters (the $3 \times 3$ rotation matrix and $3 \times 1$ translation vector) of the camera, these all are known due to pre-calibration.
In Fig. \ref{fig:back_proj}(b), the algorithm tests whether a given 3D point lies in the polygon $\{p\}_{j}$, by using the ray-casting method reported in \cite{o1998computational}.
Fig. \ref{fig:back_proj}(c) depicts the segmented 3D facial model (in the form of point cloud).


\section{CONTROL SYSTEM}    \label{sec:contr_sys}

\subsection{Path Planning}
To plan the paths for each segment a sampling-based path planning algorithm is devised.
This path planning algorithm takes point cloud of a surface as an input and outputs a set of vertical or horizontal paths (in the form of point cloud).
These paths are objected to filling the input surface with laser shots (according to the diameter of laser $\diameter_{l}$).
The path can be planned vertically or horizontally, depending on the longest side of the operating segment. 
For example, the forehead is a wider segment so horizontal paths can decrease the number of the lane changes for a robot manipulator.
Similarly, the nose region has a longer length than the width, in this case, vertical paths are more appropriate.

\begin{figure}
	\centering
	\includegraphics{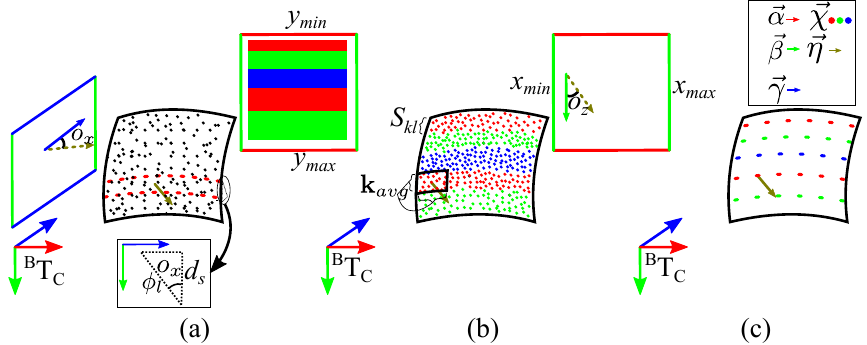}
	\caption{(on left) Partition of point cloud along $y$-axis. $\diameter_{l}$, laser diameter. $d_{s}$, laser shot separation distance. $d_{p}$, inter-path distance. $\mathbf{k}_{avg}$ averaging kernel. (on right) $\mathbf{k}_{avg}$ is incrementing with $d_{s}$ through a strip of points.}
	\label{fig:path_planning}
\end{figure}

Fig \ref{fig:path_planning} illustrates the path planning algorithm in three steps.
Initially, the algorithm bins all the point inside the range of $[y_{min}, y_{min} + \diameter_{l})$ into a container, known as strip $S_{jk}$, where the $k$ is the strip index and $j$ the segment index.
In the Fig \ref{fig:path_planning}(a), $y_{max}$ and $y_{min}$ are the maximum and minimum values along the $y$-axis among all the points in a segment.
When the surface is facing towards the camera and the $z$-axis of the camera and the normal inward to the surface are parallel.
The the range $[y_{min}, y_{min} + \diameter_{l})$ provides equals width strips to the laser diameter.
But if the surface is tilted or panned then the binning of point to a strip using the range $[y_{min}, y_{min} + \diameter_{l})$ will yield the strip's width greater than the laser diameter $\diameter_{l}$.
Which ultimately introduce the gaps among the planned paths, where the surface is curved or not facing the camera.
As only two path planning (horizontal or vertical) approaches have considered, thus rotation of the strip around $x$- or $y$-axis has computed.
In Fig. \ref{fig:path_planning}, $o_{x}$ represents the rotation of a strip around the $x$-axis, which is computed by a dot product of the projection of cumulative normal vector and $z$-axis of the camera frame ${}^{B}T_{C}$ on its $yz$-plane.
\begin{equation}    \label{eq:o_x}
    o_{x} = \cos^{-1}\left( \cfrac{(\vec{\eta}_{s}-(\vec{\eta}_{s}.\vec{\gamma}_{C})\vec{\gamma}_{C}).\ \vec{\eta}_{s,yz}}{\norm{\vec{\eta}_{s}-((\vec{\eta}_{s}.\vec{\gamma}_{C})\vec{\gamma}_{C})}\ \norm{\vec{\eta}_{s,yz}||}} \right)
\end{equation}
where the $\vec{\eta}_{s,yz}$ denoted the projected vector of cumulative normal vector of a strip $\vec{\eta}_{s}$ on the $yz$-plane of ${}^{B}T_{C}$ and $\vec{\eta}_{s,yz} = \vec{\gamma}_{C} \times \vec{\eta}_{s}$.
The $\vec{\gamma}_{c}$ points toward the $z$-axis of ${}^{B}T_{C}$.
Now the incremental step $d_{s} = \diameter_{l} / \cos(o_{x})$ will be adjusted according to $o_{x}$ and the strip $S_{kl}$ will be resorted with the new range $[y_{min}, y_{min} + d_{s})$.
All the remaining points in a segment will be binned accordingly. 
The maximum range of the previous strip will become the minimum range of the next strip as well as the maximum range of next strip will be incremented with $d_{s}$.


The width of each strip $S_{jk}$ is equal to the diameter of laser shots $\diameter_{l}$, so a line passing through the centre of each strip can be considered as an optimal path for the robot end-effector.
Optimal path in the sense that if the robot manipulator follows it, no overlapping of laser shots would occur among the paths of a segment.
This path can be computed using a polynomial fit, but this method can not grantee that the fitted polynomial will pass from the centre of a strip of points.
Furthermore, the points in each strip are sparsely distributed which can also bias the polynomial fit.
Then the fitted polynomial could generate a path deviated from the centre of the strip.
That is why this method has been devised to ensure that the planned paths always pass through the centre of each strip.

Once the strip $S_{jk}$ is partitioned from a segment of point cloud, a patch $\mathbf{k}_{avg}$ (like a rectangular stencil) of width $\diameter_{l}$ and height $d_{s}$ is placed on the one side of the strip, shown in Fig. \ref{fig:path_planning}(b).
The patch $\mathbf{k}_{avg}$ is a buffer of the points, which populates the points after every incremental/decremental step and compute the average of all enclosed points inside it.
The size of the step is equal to the diameter of laser $\diameter_{l}$.
Mathematically, $\mathbf{k}_{avg}$ can also be defined as a set of points:
\begin{equation}
\mathbf{k}_{avg} = \{s_{1}, \dots, s_{i}\} 
\end{equation}
where $s_{i} = \{\vec{x}_{i}, \vec{n}_{i}, \vec{c}_{i}\}$ and $\vec{x}_{i} = [x_{i}\ y_{i}\ z_{i}]^{T}$.
$s_{i} \in \mathbf{k}_{avg}$ $\iff$ $(x_{\mathbf{k}_{avg}} - d_{s}/2) < x_{i} < (x_{\mathbf{k}_{avg}} + d_{s}/2)$, where $x_{\mathbf{k}_{avg}}$ is the position of the patch $\mathbf{k}_{avg}$ in the strip.
In Fig. {\ref{fig:path_planning}}, a patch $\mathbf{k}_{avg}$, in second strip from the top, is sweeping from left to right with $d_{s}$ incremental step.
The position of the patch $\mathbf{k}_{avg}$ in a strip is defined as:\\
\begin{equation}
x_{\mathbf{k}_{avg}} = x_{min} + a  d_{s} \quad, \quad a = 0,1,2\dots, L
\end{equation}
where $L$ is the length of the strip and is defined as $L=\text{int}((x_{min} - x_{max})/d_{s})$.
$x_{min}$ and $x_{max}$ are the minimum and maximum values along $x$-axis of a strip $S_{jk}$.

\begin{figure}
	\includegraphics{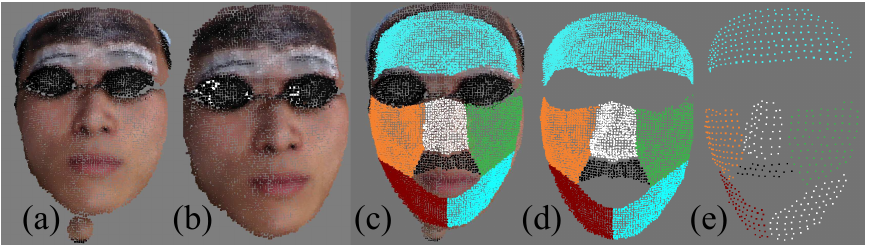}
	\caption{Facial Reconstruction to Path Planning. (a) raw facial model, obtained from scanning. (b) clean facial model, unnecessary point discarded using GUI. (c) and (d) 3D segmentation. (e) planned path for each region}
	\label{fig:segmentation}
\end{figure}

Now, the $P_{j}$ are the paths for a segment $S_{j}$ and a path point $p_{i}^{j}$ is a structure of vector and defined as
\begin{equation}
p_{i}^{j} = \{ \vec{\chi}_{i}^{j}, \vec{\eta}_{i}^{j} \}
\end{equation}
$\vec{\chi}$ and $\vec{\eta}$ are the position and normal vectors of $3 \times 1$.
These two entities obtain from the average of position and normal vector of the points inside the patch $\mathbf{k}_{avg}$:
\begin{align}
\vec{\chi} = \frac{1}{N_{k}} \Sigma_{i}^{N_{k}} \vec{x}_{i} , \qquad
\vec{\eta} = {\frac{1}{N_{k}} \Sigma_{i}^{N_{k}} \vec{n}_{i}}/{\norm{\frac{1}{N_{k}} \Sigma_{i}^{N_{k}} \vec{n}_{i}}}
\end{align}
$N_{k}$ is the number of points inside the patch.
The patch $\mathbf{k}_{avg}$ moves from the $x_{min}$ to $x_{max}$ in the first strip and then return from $x_{max}$ to $x_{min}$. 
The back and forth motion of the patch in the adjacent strips enforces the algorithm to generate an S-shaped path for a segment.
Fig. {\ref{fig:segmentation}} illustrates the output from each of the proposed algorithms (scanning to path planning).

\subsection{Robot Control Framework}
\begin{figure}
	\centering
	\includegraphics[width=\columnwidth]{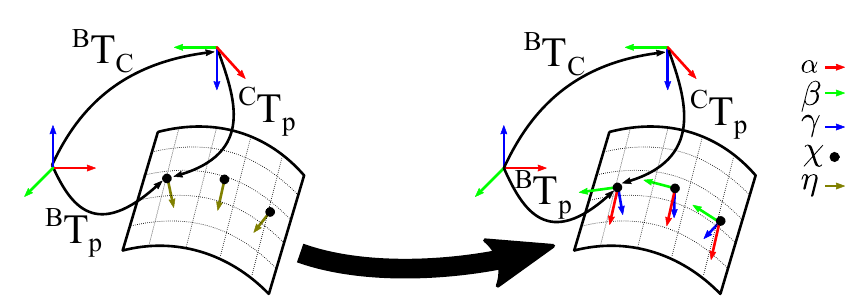}
	\caption{Calculation of three column vectors $\vec{\alpha}_{p}$, $\vec{\beta}_{p}$ and $\vec{\gamma}_{p}$ of a rotation matrix ${}^{B}R_{p}$ from a surface normal vectors $\vec{\eta}$.}
	\label{fig:normal_vector_to_rot_mat}
\end{figure}

\begin{figure}
    \centering
    \includegraphics{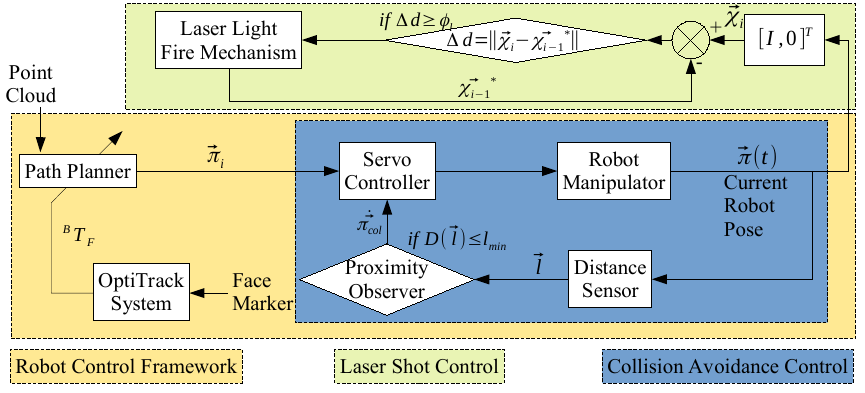}
    \caption{Three proposed control systems to guarantee a safe and accurate robotic photo-rejuvenation.}
    \label{fig:control_systems}
\end{figure}

Each point in a planned path is a structure of two vectors $\vec{\chi}$ and $\vec{\eta}$.
For this dermatological procedure, the laser light source should project the laser light normal to the skin surface at the instance of delivering laser light energy.
This can maximise the energy transfer to the skin in each laser shot.
$\vec{\chi}$ and $\vec{\eta}$ are converted into a position command for a robot manipulator.
$\vec{\chi}$ can directly define the desired position of robot end-effector. 
To define the orientation of robot end-effector, first, the rotation matrix derives from the normal vector. 
As a rotation matrix is composed of three column vectors and each vector is orthonormal to others, i.e. ${}^{B}R_{p} = [\vec{\alpha}_{p}, \vec{\beta}_{p}, \vec{\gamma}_{p}]^{T}$.
Let's consider that the $\vec{\gamma}_{p}$ and $\vec{\eta}$ are parallel normal vectors.
Then the cross product of the second column of rotation matrix of robot base $\vec{\beta}_{B}=[0\ 1\ 0]^{T}$ and the third column  $\vec{\gamma}_{p}$ of desired rotation matrix ${}^{B}R_{p}$ at a path point gives  $\vec{\alpha}_{p} = \vec{\beta}_{B} \times \vec{\gamma}_{p}$.
$\vec{\beta}_{p}$ can be computed as $\vec{\beta}_{p} = \vec{\gamma}_{p} \times \vec{\alpha}_{p}$.
Fig. {\ref{fig:normal_vector_to_rot_mat}} provides a better illustration of the normal vectors calculation of rotation matrix from a surface normal vector.
The robot manipulator used in the proposed prototype can only recognise the rotation in the axis-angle representation.
Suppose an \st{unit} arbitrary vector $\vec{u}$ is parallel to the rotation axis of a rigid body rotation defined by a rotation matrix ${}^{B}R_{p}$ of $3\times3$.
\begin{equation}
{}^{B}R_{p} = \left[ \vec{\alpha}_{p}\ \vec{\beta}_{p}\ \vec{\gamma}_{p} \right] = 
\begin{bmatrix}
\alpha_{px}        &    \beta_{px}        &    \gamma_{px}\\
\alpha_{py}        &    \beta_{py}        &    \gamma_{py}\\
\alpha_{pz}        &    \beta_{pz}        &    \gamma_{pz}
\end{bmatrix}
\end{equation}
Then the vector $\vec{u}$ is calculated as \cite{baker2012matrix}
\begin{equation}
\vec{u} = 
\begin{bmatrix}
\beta_{pz} - \gamma_{py}\\
\gamma_{px} - \alpha_{pz}\\
\alpha_{py} - \beta_{px}
\end{bmatrix}
\end{equation}
The magnitude of $\vec{u}$ is equal to $||\vec{u}||=2sin\theta$ and $\theta$ is the angle to which a rigid body rotates around $\vec{u}$ {\cite{baker2012matrix}}.
Where $\theta$ is computed as
\begin{equation}
|\theta| = \cos^{-1} \left(\cfrac{\Tr({}^{B}R_{p}) - 1}{2}\right).
\end{equation}
Now the representation of rotation of a rigid body in axis angle is
\begin{equation}
\nu = \theta \vec{u}
\end{equation} 
To define the pose of the robot end-effector for each path point, the pose vector will be;
\begin{equation}
\vec{\pi}_{i} = [\vec{\chi}_{i}^{T} \quad \vec{\nu}_{i}^{T}]^{T}
\end{equation}
where $\vec{\pi}_{i}$ is a pose vector of $6 \times 1$ for each path point $p_{i}$.
The generated paths are subjected to update in the case of noticeable motion of the human head.
Fig. \ref{fig:control_systems} shows all of the control systems and their inter-connection with other control systems as well. 

The maximum velocity depends on the diameter of the laser $\diameter_{l}$ and the pulse rate of the laser machine $\tau_{l}$.
\begin{equation}
    \norm{\dot{\vec{\pi}}_{max}} = \diameter_{l} \tau_{l}
\end{equation}
This consideration ensures the continuous motion of the robot end-effector while delivering laser energy, instead of stop the robot, deliver the laser light energy and move to next spot. 

\subsection{Laser Shot Control}
To deliver the laser energy uniformly over the treating skin surface, the control of the laser firing instance is paramount.
To define a laser-firing instance, an impulse function $\delta_{i}(\Delta d - \diameter_{l})$ is implied. 
Where the $\Delta d$ is the argument of the impulse function and is defined as the distance covered by the robot end-effector from the position of last laser shot $\vec{\chi}_{i-1}^{*}$ to current position $\vec{\chi}_{i}$, so $\Delta d = \norm{\vec{\chi}_{i} - \vec{\chi}_{i-1}^{*}}$. 
Where $\diameter_{l}$ defines the diameter of a laser shot.
This impulse function generates $1$ as an output only when its argument becomes zeros.
Then the instance of laser shot can be defined as,
\begin{equation}\label{eq:instance}
\delta_{i}(\Delta d - \diameter_{l}) = 
\begin{cases}
1 &\Delta d \geq \diameter_{l}\\
0 &\text{else}
\end{cases}
\end{equation}
The laser shot instance only occurs when the robot is following the pose vectors $\pi_{p_{i}}^{B}$.
Otherwise, the output from $\delta_{i}(.)$ will not consider.
So the pose vector of the end-effector when the laser shot occurs can be defined as,
\begin{equation}\label{eq:instance_vec}
\vec{\psi}_{i}\ =\ \delta_{i} \vec{\pi}_{i}
\end{equation}
$\vec{\psi}_{i}$ denotes the pose vector of $6 \times 1$ of end-effector where the laser shot instance occurred.
This laser fire control relies on $\Delta d$, where the $\Delta d$ depends only on the current position of robot end-effector and the position of last laser shot. 
After every laser shot instance, the $\Delta d$ is reset to zero.
This simple control of laser shot performs elegantly to distribute the laser shot uniformly. 



\subsection{Collision Avoidance Control}    \label{ssec:collision_avoidance_control}
It is necessary for the robot end-effector to maintain a safe distance to the human face while performing rejuvenation treatment.
To equip the robotic system with the proximity-like behaviour, three distance sensors are housed in the end-effector besides the soft cap, as shown in Fig \ref{sec:pros_rob_sys}.
This ensures that the system is aware of end-effector's surrounding and can avoid any physical contact between end-effector and the face during the treatment. 
These point distance sensors can only provide the distance from the sensor's receiver to reflecting surface.
To reliably include the distance measurement provided by these sensors in the system, we are utilising their accumulated weighted averaged, using,
\begin{equation}
    \vec{l} = \frac{\Sigma_{m} \omega_{m} \vec{l}_{m}}{\Sigma_{m} \omega_{m}} \quad , \quad \omega_{m} = \frac{\vec{l}_{m} . \vec{\eta}_{*}}{\norm{\vec{l}_{m}}\norm{\vec{\eta}_{*}}}
\end{equation}
Where, $\vec{l}$ is the weighted averaged vector from the sensors to the surface, $\omega_{m}$ the weight based on the dot product of the two vectors, $\vec{l}_{m}$ the vector parallel to the measuring direction of the distance sensor, $\vec{\eta}_{*}$ the normal vector pointing inward to the surface of the face and located at the intersection of $\vec{l}_{m}$ and the surface of the facial model.
$m$ represent the index  number of each distance sensor.

\begin{figure}
    \centering
    \includegraphics{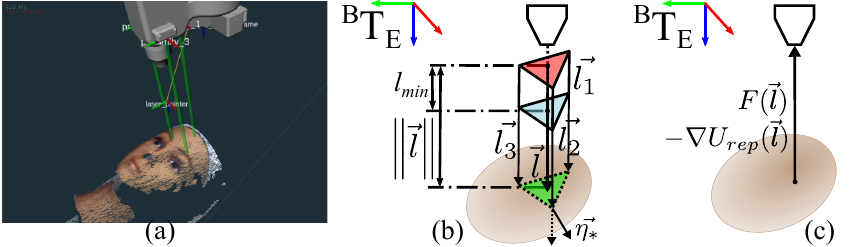}
    \caption{Collision Avoidance Control. (a) Simulated realisation of the method. (b) Distance vectors. (c) Force acted on the robot end-effector.}
    \label{fig:distance_sensing}
\end{figure}

To implement a repulsive behaviour in order avoid the collision between the face and end-effector, a controller in the back-end is continuously observing the $\vec{l}$ and intervenes, only when $\norm{\vec{l}} < l_{min}$, where the $l_{min}$ is the least close allowed distance for robot end-effector.
This behaviour is the same as two repelling bodies in an artificial potential field.
In this context, if we can consider the robot end-effector as a moving agent and the facial surface as a boundary and there is  a repulsive force between them, which is defined by the gradient of the artificial potential function as in \cite{choset2005principles}, i.e, 
\begin{equation}
    U_{rep}(\vec{l}) = \begin{cases}
    \frac{1}{2}\left( \frac{1}{l_{min}} - \frac{1}{D(\vec{l})} \right)^{2} & D(\vec{l}) \leq l_{min}    \\
    0 & D(\vec{l}) > l_{min}
    \end{cases}
\end{equation}
and the gradient will be
\begin{equation}
    \nabla U_{rep}(\vec{l}) =  \begin{cases}
    \left( \frac{1}{l_{min}} - \frac{1}{D(\vec{l})} \right)\frac{1}{D(\vec{l})^{2}} \Delta D(\vec{l}) & D(\vec{l}) \leq l_{min}    \\
    0 & D(\vec{l}) > l_{min}
    \end{cases}
\end{equation}
Here $U_{rep}(l)$ is the potential field between robot end-effector and facial model whereas $D(\vec{l})$ represent the distance function of $l$, as shown in Fig \ref{fig:distance_sensing}.
The artificial force acted on the robot end-effector will be, $F(l) = -\nabla U_{rep}(\vec{l})$ and the generated velocities to push the end-effector away from the facial surface the resulting velocities will be $\dot{\vec{\pi}}_{col} = \kappa F(\vec{l})$, where $\kappa$ is the arbitrary scaling factor. 

\subsection{User Control Interface}
A user control interface is developed for an end-user (physician and practitioner).
In this user control interface, a reconstructed facial model is visualised for user interaction.
One can discard the unnecessary or floating points if present in the facial model before feeding the model to the automatic face segmentation algorithm.
It will ensure that the noisy points in the model do not affect the paths planned from the segments.

The user control interface allows the user to initialise the procedure for any segmented region and a selection tool is also provided in the case user wants to operate on a custom-defined region of skin.

\section{RESULTS}\label{sec:results}

\subsection{Cross-Calibration}
Cross-calibration is a method to validate a transformation link or a coordinate frame from multiple transformation chains (or kinematic chains).
The proposed system is relying on multiple sensors and actuators to perform the complete rejuvenation task.
Each sensor is measuring in its own coordinate frame, e.g. the depth camera, distance sensors and tracking system. 
Likewise, the actuators i.e. robot manipulator and laser light generator (handpiece) require the control command in their own frame of reference.
Thus, without the accurate knowledge of the relationship between each coordinate frame, the complete proposed system is not able to perform the treatment properly.

\begin{figure}
	\centering
	\includegraphics[width=\columnwidth]{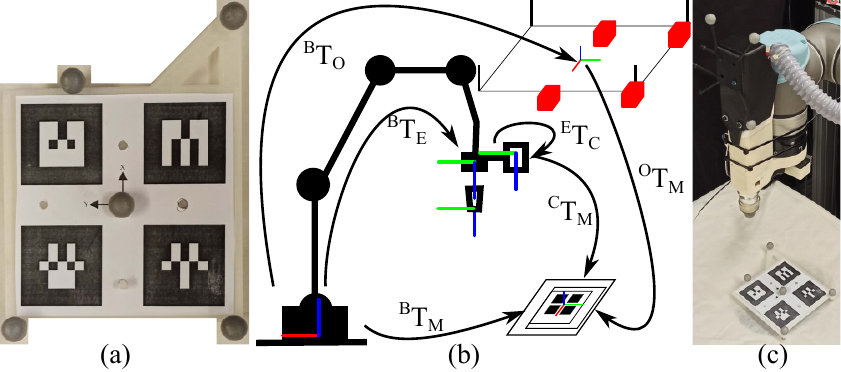}
	\caption{Setup to acquire and validate the transformations between different coordinate frames. (a) AR/OptiTrack Marker. (b) conceptual illustration of important transformation frames. (c) physical illustration. }
	\label{fig:cali_valida_opti}
\end{figure}

Fig \ref{fig:cali_valida_opti} illustrates the physical setup to conduct cross-calibration.
A marker, as shown in Fig. \ref{fig:cali_valida_opti} (a), is composed of AR markers and tracking markers and has a common transformation frame for camera and tracking system. 
When the tracking system and depth camera is observing this marker at a same time then a transformation from robot base to tracking system's origin can be found, ${}^{B}T_{O} = {}^{B}T_{E} {}^{E}T_{C} {}^{C}T_{M} ({}^{O}T_{M})^{-1}$.
Where, the different superscripts and subscripts are corresponding as,  $B$ the robot base, $O$ the OptiTrack motion tracking system, $E$ the end-effector, $C$ the depth camera and $M$ the AR/OptiTrack marker.
The subscript and subscript represent the parent and child transformation frame.  
Once the ${}^{B}T_{O}$ is known, any transformation frame can be computed or validated.

\subsection{Laser Shots Separation Distance Test}
\begin{figure}
	\centering
	\includegraphics[width=\columnwidth]{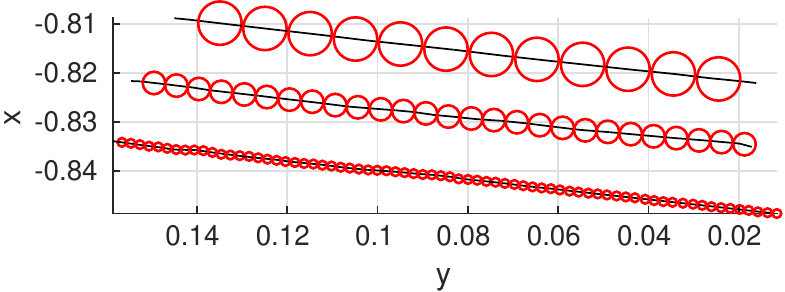}
	\caption{Laser shot separation 0.01m, 0.005m and 0.002m (top to bottom)}
	\label{fig:traj_valida_plot}
\end{figure}

\begin{figure}
	\centering
	\includegraphics[width=\columnwidth]{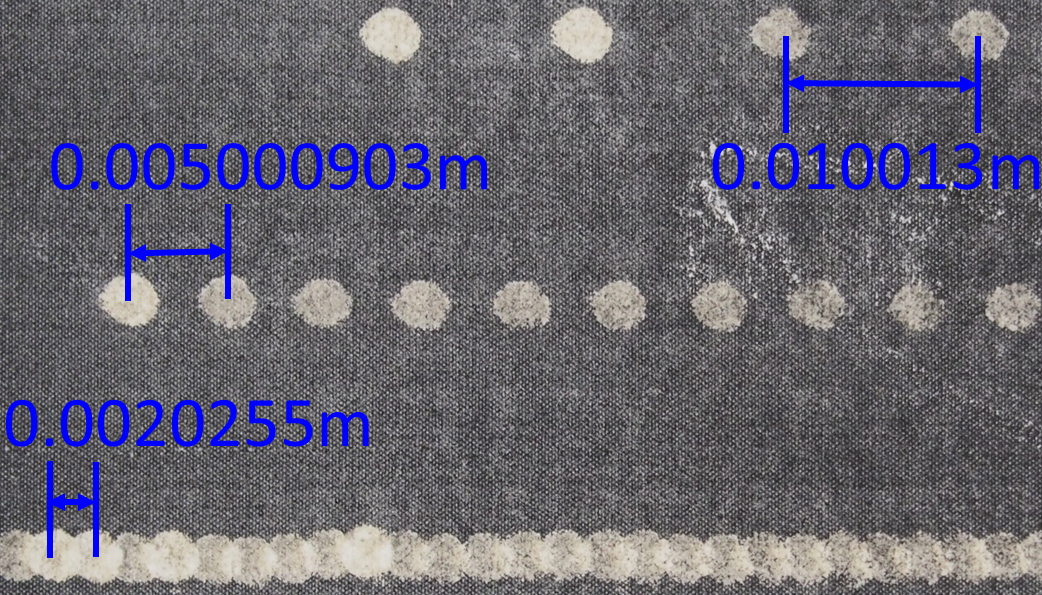}
	\caption{Laser shot separation distance test: the values shown in blue colour are the measurement from vernier calibre.}
	\label{fig:traj_valida}
\end{figure}

The laser shot control is evaluated by the separation distance test.
In this test, a paper printed with the black ink is placed on cardboard and let the robot end-effector followed a linear path while firing the laser shot of 3mm diameter on it.
For each run, different laser separation distance $d_{s}$ was assigned, $d_{s} = 0.01m,\ 0.005m\ 0.002m$.
The pose of end-effector when the laser shot instance occurred was recorded, using Eq. \eqref{eq:instance_vec}.
In Fig. \ref{fig:traj_valida_plot}, these laser shot instances are plotted with red circles and the path followed by the robot end-effector with a black line.
Furthermore, the distance between each shot is also measured by vernier calibre to validate the measurements, as shown in Fig. \ref{fig:traj_valida}.
Table \ref{tab:sep_dist} shows that in each test the laser firing control has repeatedly trigger the laser shot instance precisely at the desired distance.
That is why the mean value of each test is near to the desired distance and the variance value is infinitesimally small. 
The highest variance is recorded for SDT3 where the pre-defined laser diameter $\diameter_{l}$ is 2mm.
This can be improved by decreasing the end-effector's speed which provides the system with a wider window of time to acquire the end-effector position from robot manipulator.
The last column of Table. \ref{tab:sep_dist} illustrates the area covered by the laser shots.

\begin{table}
	\begin{center}
		\caption{Evaluation of separation distance consistency} \label{tab:sep_dist}
		\begin{tabular}{lcccccc}
			\hline
			Tests &$N_{t}$   &d(m) 	&$\diameter_{l}$(m)    &$\mu$(m)  &$\sigma^{2}$(nm) & $\Phi$(\%) \\ 
			\hline
			SDT1  &11  &0.111   &0.01                  & 0.01009  & 2.09            & 77.832\\  
			SDT2  &26  &0.131   &0.005                 & 0.00505  & 2.11            & 77.940\\  
			SDT3  &71  &0.146   &0.002                 & 0.00205  & 4.05            & 76.388\\  
			\hline
		\end{tabular}
	\end{center}
	\footnotesize{SDT: separation distance test. $N_{t}$: number of laser shots fired in each test. $\diameter_{l}$: radius of the laser. d: Distance covered by the end-effector. $\mu$: mean distance of the laser shots. $\sigma^{2}$: variance in the laser shots. $\Phi$: area covered by the laser shots.}\\
\end{table}

\subsection{Real-time Planned Path Update}
\begin{figure*}[ht]
    \centering
    \includegraphics[width=\textwidth]{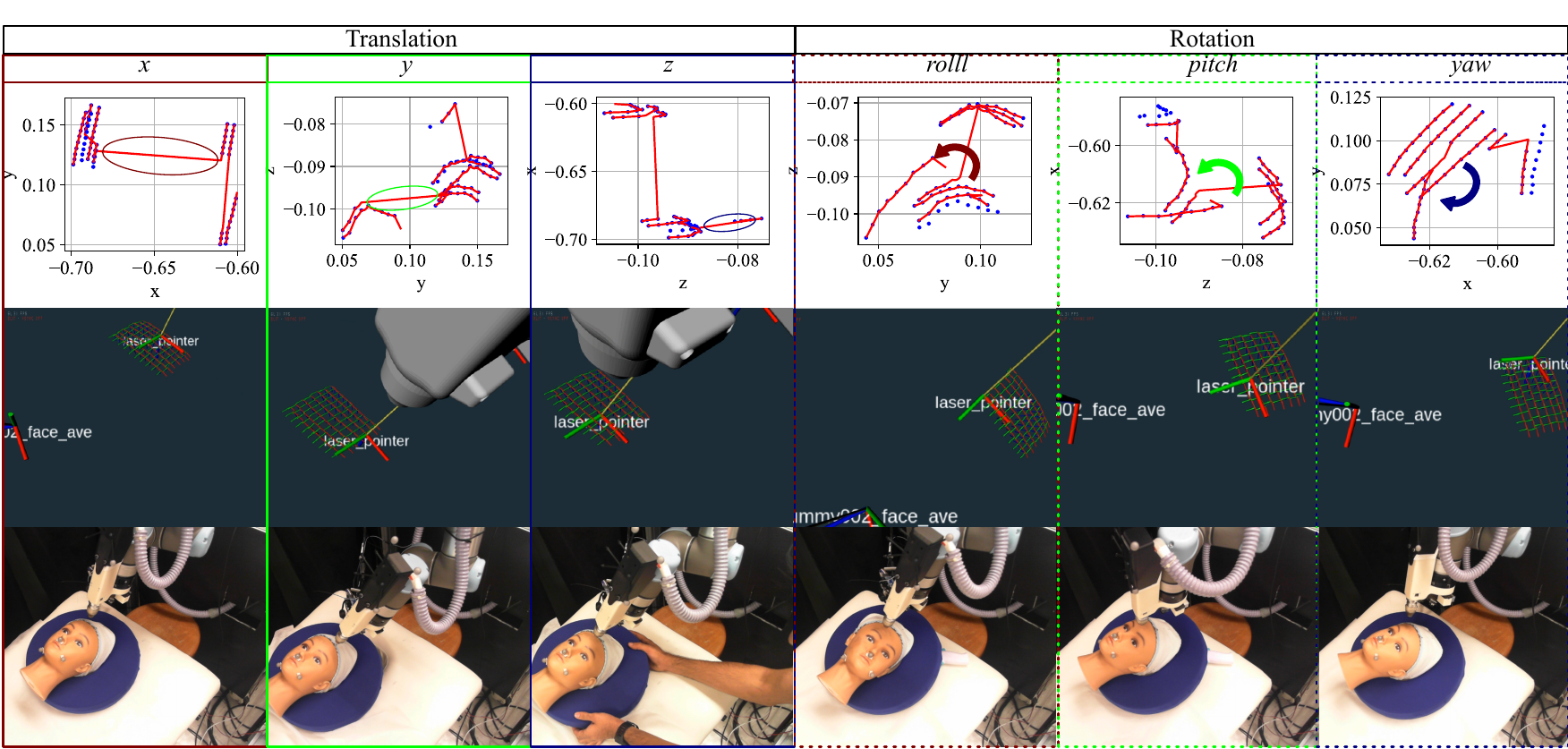}
    \caption{Real-time adjustment of the planned paths and robot manipulator motion while performing rejuvenation procedure. }
    \label{fig:tracking_result}
\end{figure*}

Fig. \ref{fig:tracking_result} demonstrates the real-time response of the system in the case of sudden movement of the human head during the treatment.
Three markers have attached to the face at upper lips, left and right jaw to continue tracking the pose of the face.
The system ignores the small linear and angular motion within $3mm$ and $4^{o}$ respectively.
In the first row of Fig. \ref{fig:tracking_result}, the blue dots depict the desired paths and the red shows the followed path by the end-effector. 
The second row illustrates the path updates and follows in parallel real-time simulation. 
Whereas, the motion of the mannequin's head can be seen in the pictures in the third row. 
The cherry red, green, and navy blue ellipses in each of the first three graphs in the second row represent the linear motion in $x$, $y$ and $z$-axis, respectively. 
Similarly, the cherry red, green, and navy blue arrows display the rotation in $x$, $y$ and $z$-axis, respectively (roll, pitch and yaw).

\subsection{Collision Avoidance Test}
\begin{figure}
    \centering
    \includegraphics{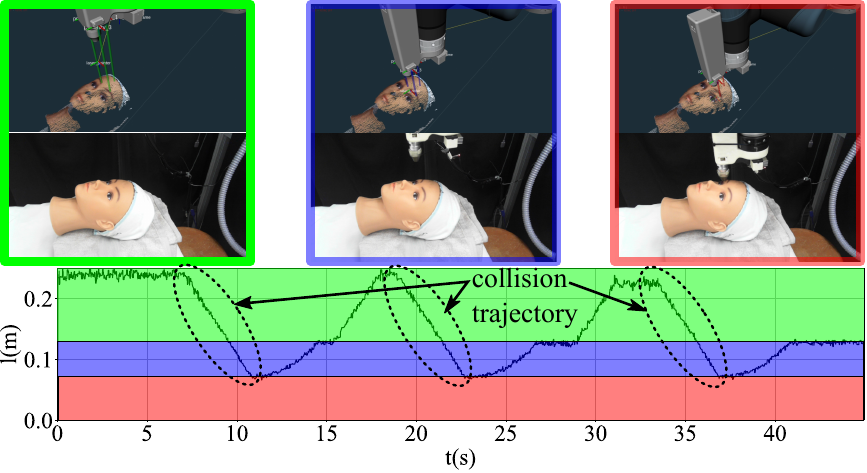}
    \caption{Collision avoidance control test. Green colour represents the free zone, blue the operation zone and the red danger zone. First and second rows illustrates the simulated and real world setting for the test. Whereas the third row is a graph of distance $\norm{\vec{l}}$ versus time $t$. }
    \label{fig:collision_avoidance_control_test}
\end{figure}

To assess the collision avoidance control proposed in the \ref{ssec:collision_avoidance_control}, the robot end-effector is enforced to move dangerously near to the mannequin face (to mimic the possible scenario). 
Fig. \ref{fig:collision_avoidance_control_test} illustrates the three levels of closeness according to the desired behaviour.
In Fig. \ref{fig:collision_avoidance_control_test}, the green colour represents the ``free zone'', the blue colour the ``operation zone'' (the distance required to perform the procedure) and the red colour the ``danger zone'' (the closeness can cause a collision between end-effector and a face)

Fig \ref{fig:collision_avoidance_control_test} illustrates the implementation of the collision avoidance control.
In this test, we have provided the robot end-effector with a collision trajectory three times. 
The distance from the robot end-effector and the face $l$ versus time $t$ is plotted in the third row of the Fig. \ref{fig:collision_avoidance_control_test}.
It is evident that the controller is preventing the robot end-effector from entering the red zone and pushing the end-effector back to the operation zone.

\
\subsection{Test on the Human Subjects}
\begin{figure}
    \centering
    \includegraphics{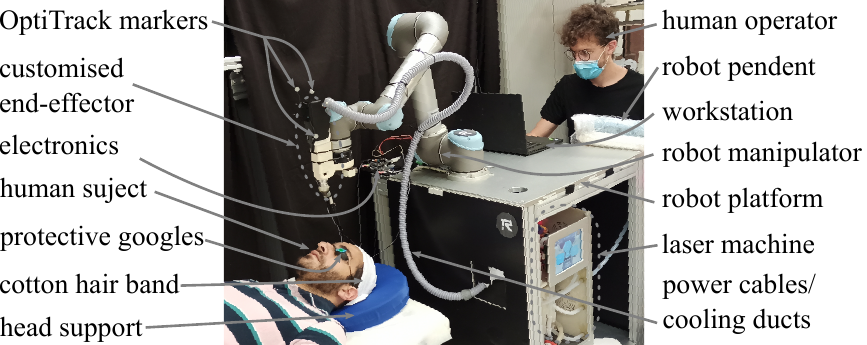}
    \caption{The proposed robotic prototype and its components.}
    \label{fig:physical_setting}
\end{figure}

Fig \ref{fig:physical_setting} depicts the proposed robotic prototype.
The proposed automated skin photo-rejuvenation procedure has been performed on the human volunteers \footnote{Ethics Approval Reference Number: HSEARS20200604002, Human Subjects Ethics Sub-committee, Departmental Research Committee, The Hong Kong Polytechnics University, Hong Kong}.
Before performing the automated procedure on the human volunteers, the system has been tested repeatedly on a mannequin's face.
Although the effectiveness, accuracy, and robustness of the proposed system can be evaluated by performing the automated procedure on a mannequin's face.  
But the aesthetic outcome of the mannequin's treated face can't be visualised or predicted. 
The human volunteers wore a set of protected glasses to avoid direct contact of laser shot with their eyes or eyelids.
The eyebrows were also covered by medical paper tape.
Their heads had been covered with a strip of thick cotton cloth to avoid laser exposure to their hairs.

\begin{figure}
	\centering
	\includegraphics[width=\columnwidth]{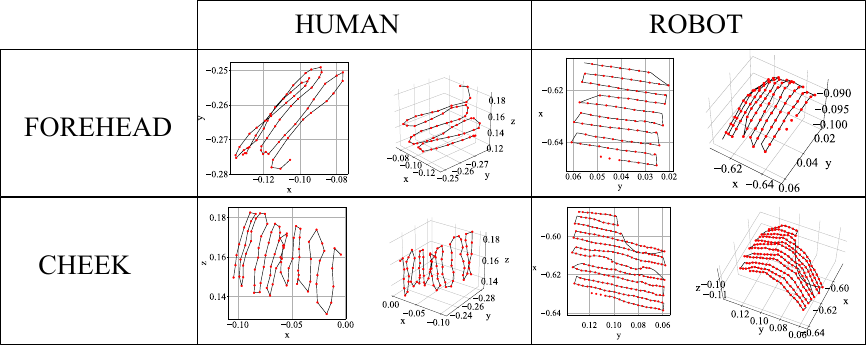}
	\caption{Motion and laser shot instance plot for both human and robot operator(units are in meters).}
	\label{fig:result_shot_fro}
\end{figure}

\begin{figure}
    \centering
    \includegraphics{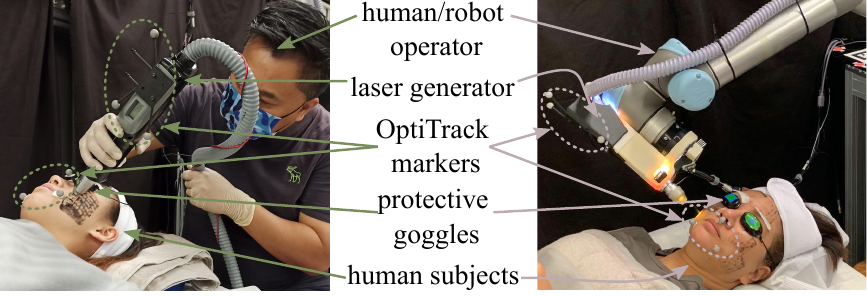}
    \caption{Skin photo-rejuvenation treatment is conducting by human and robot operator.}
    \label{fig:human_and_robot}
\end{figure}

To validate the proposed system and compare with an experienced practitioner, we drew the square patches of $47\times47$mm with charcoal on the human subject of fixed size.
This selection keeps the comparison unbiased between human and robot.
Then, let the human and robot operators perform the treatment inside drawn patches on the left and right side of the face of the human subject.
The human operator was asked to fill the charcoal patch with laser shots while avoiding overlapping the laser shots and set the same rules for the robotics prototype, as shown in Fig \ref{fig:human_and_robot}.
For the human subjects, the Q-Switch Nd: YAG laser of 1064nm wavelength with 4mm laser diameter is used and energy was set to 600$mJ$ where the fluence was 10$J/cm^{2}$.
The defined parameters were similar for both operators.
In Fig. \ref{fig:result_shot_fro}, the path followed by the robot end-effector is represented by a black line and the centre of each red circle is the instance of a laser shot. 
The diameter of each red circle is equal to the diameter of the laser $\diameter_{l}$.
The path and the laser shot instances are based on real-time data recorded while performing the procedure. 
The 3D plots are shown in Fig. \ref{fig:result_shot_fro} illustrates the effectiveness of the procedure in the term of uniform laser distribution and path consistency.

\begin{table}[]
\begin{center}
\caption{Evaluation of separation distance consistency} \label{tab:human_vs_robot}
\begin{tabular}{cccccccc}
\cline{3-8}
   &   & \multicolumn{3}{c}{HUMAN} & \multicolumn{3}{c}{ROBOT} \\ \hline
ID &
  $N_{p}$ &
  \begin{tabular}[c]{@{}c@{}}$t_{h}$\\ (s)\end{tabular} &
  $N_{h}$ &
  \begin{tabular}[c]{@{}c@{}}$\Phi_{h}$\\  (\%)\end{tabular} &
  \begin{tabular}[c]{@{}c@{}}$t_{r}$\\ (s)\end{tabular} &
  $N_{r}$ &
  \begin{tabular}[c]{@{}c@{}}$\Phi_{r}$\\  (\%)\end{tabular} \\ \hline
1  & 3 & 2040    & 323   & 61.24   & 1860    & 341   & 64.66   \\
2  & 3 & 2580    & 350   & 66.36   & 1920    & 357   & 67.69   \\
3  & 3 & 2460    & 343   & 65.04   & 2160    & 346   & 65.60   \\
3  & 2 & 1560    & 162   & 46.07   & 2040    & 245   & 69.68   \\
4  & 2 & 1080    & 153   & 43.51   & 3120    & 262   & 74.52   \\
5  & 3 & 1140    & 197   & 37.35   & 1740    & 381   & 72.24   \\
6  & 2 & 1140    & 197   & 56.03   & 1740    & 241   & 68.54   \\
7  & 4 & 4020    & 290   & 41.24   & 3780    & 523   & 74.37   \\
8  & 3 & 1740    & 259   & 49.11   & 1500    & 355   & 67.31   \\
9  & 2 & 1560    & 172   & 48.92   & 1860    & 260   & 73.95   \\
10 & 2 & 1380    & 160   & 45.50   & 2340    & 239   & 67.98   \\
11 & 2 & 2520    & 155   & 44.08   & 5100    & 274   & 77.93   \\ \hline
   &   &         &       &         &         &       &        
\end{tabular}
\end{center}
\footnotesize{$N_{p}$: number of charcoal patches. $t_{h}$ and $t_{r}$ are time taken by a human and robot operator, respectivley. Similarly, $N_{h}$ and $N_{r}$ the number of laser shots, $\Phi_{h}$ and $\Phi_{r}$ the area covered by the laser shots.}\\
\end{table}

\begin{figure}
    \centering
    \includegraphics{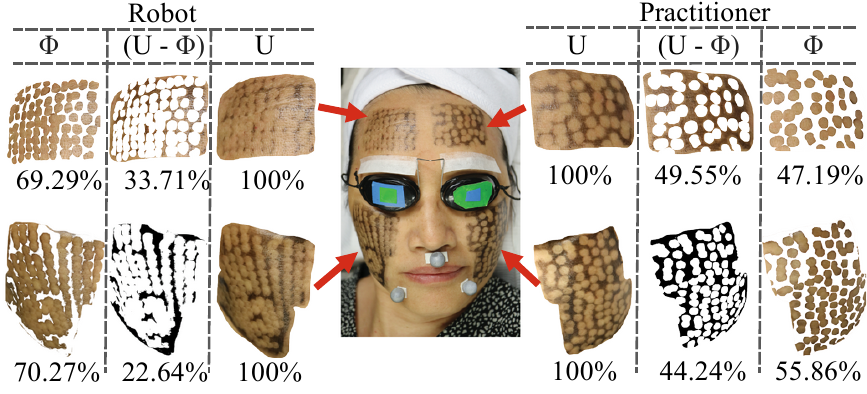}
    \caption{Comparison of covered area $\Phi$ in the oeprable area $U$.}
    \label{fig:FACE_CHARCOAL_VS_TREATED_NoPixels}
\end{figure}

In Table. \ref{tab:human_vs_robot}, the comparison of the operable area covered by the human and robot operator is shown. 
Table. \ref{tab:human_vs_robot} shows the comparison of the operable area covered by the human and robot operator.
According to this, the mean and the standard deviation of the area covered $\Phi_{h}$ by the human operator is 50.37\% and 9.56\%, respectively.
Whereas, the robotic system has achieved 70\% of the mean in area coverage $\Phi_{r}$ with the standard deviation of 4.13\%. 
Which is about 20\% more area coverage and half the standard deviation in it than the human operator.
Here one fact is worth to mention that \emph{an inscribed circle in a square} can cover only 78.54\% of the total area of the circle.
Now, if we compare the value in the column under $\Phi_{r}$, it provides a better comparison of covered vs operable area.
Fig. \ref{fig:FACE_CHARCOAL_VS_TREATED_NoPixels} depicts the covered vs. operable region, here operable region is denoted by $U$ and covered with $\Phi$.

\begin{figure}
    \centering
    \includegraphics{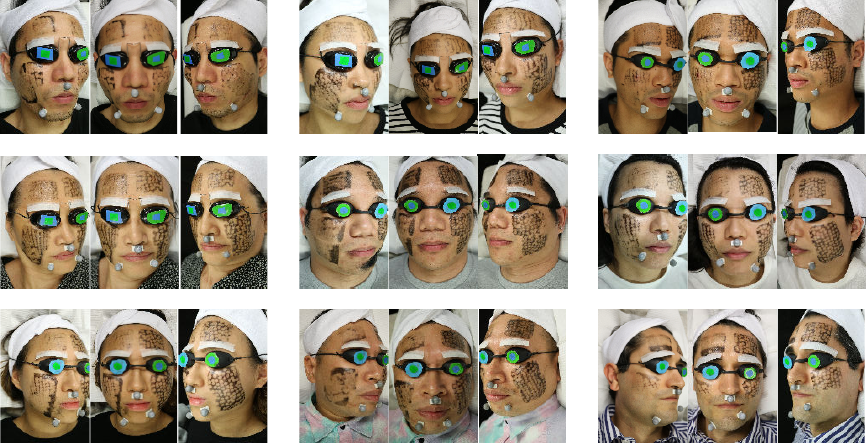}
    \caption{Human Subjects. The robotic system has performed the skin photo-rejuvenation on the left-hand side of each human subject, while an experienced human operator has perform on the right-hand side.}
    \label{fig:collage}
\end{figure}

Fig. \ref{fig:human_and_robot} shows the physical setup to conduction human trails.
Both operators are equipped with the same laser generator from the same company.
The motion of the laser handpiece during the treatment has recorded using the OptiTrack motion tracking system, for both operators.
The path followed by the human operator and the robot can be seen in Fig. \ref{fig:result_shot_fro}.
Fig. \ref{fig:collage} provides a visual comparison of the area coverage without overlapping.
The validation and the demonstration of the proposed algorithm can been seen from \url{https://github.com/romi-lab/cosmetic_robotics/raw/main/videos/video.mp4}.

\section{CONCLUSIONS}\label{sec:conclusions}
In this article, we have demonstrated the automated facial skin rejuvenation robotic prototype.
To complete the manipulation task, an industrial-grade robotic manipulator along with the custom end-effector was used.
The custom end-effector equipped with a depth camera, distance sensors and laser generator.
A method was presented to estimate the pose of the human face and the viewpoint around the face.
An accurate facial model was constructed from the point cloud data captured from the different estimated viewpoints.
The 3D facial model was segmented into seven regions using the 2D facial landmarks.
Then the optimal paths of robot manipulator extracted from each segmented.
The optimal path ensured that no overlapping of laser shots between the paths occurred.
To control the distance of laser shots while following the optimal path, a control law was devised to fire the laser shot after predefined laser separation distance.
The proximity like behaviour has been implemented to avoid any possible collision between robot end-effector and the human face. 
Initially, the facial model is reconstructed in the camera coordinate and the motion of the face is tracked through another system beside this the robot needs position commands in its own coordinate frame.
To avoid the possible error could be introduced by a relation of each transformation frame, a cross-calibration technique is utilised.
This error reduction is achieved by observing the same AR/OptiTrack marker ${}^{B}T_{M}$ from two transformation chains then minimise the difference between them.
The laser separation distance is also evaluated by firing the laser shot on a plain piece of paper.
Then the separation of laser shots is also measured using vernier calibre.
The laser shots were plotted on a graph to check the possible overlapping of laser shots.
The validation of each proposed control system has been covered thoroughly.

The uniform laser distribution demonstrates the potential improvement in the outcomes of this procedure, which can not be achieved by a human operator.
But, the proposed methods have some limitations.
As the laser fire controller is only using the local information of two positions, $\vec{\chi}_{i}$ and $\vec{\chi}_{i-1}^{*}$  which could cause the overlapping of laser shots at the edges of the paths or the path changing.
At the current stage, the system cannot check or adjust the laser energy level, which needs a supervisory role from the operator to adjust the laser energy to the desired level. 
Future works also include the fusion of current skin tissue temperature in the control of laser firing.
This will enable the robotics system to decide whether to fire or not to fire on an instance.
Furthermore, it is also observed that only temperature measurement of the surface of the skin may not be sufficient to develop a laser fire controller.
Because the temperature of the skin surface can be measure by a thermal camera and these measurements depend on the emittance $\varepsilon$ of the observed object, each temperature measurement may perturb from a real value \cite{flirets3xx}.
The room temperature also influences the temperature of the skin surface.
Another reason is a low thermal diffusivity of skin outer layer ``stratum corneum'', which makes the outer layer of the skin a good insulator.
Thus, the temperature changes occurred inside the skin tissue due to laser-skin interaction cannot be predicted by the temperature values on the skin surface.
In-depth clinical study will be conducted to compare the efficiency of skin photo-rejuvenation procedure by the proposed robotic prototype and a human operator.
Not only to compare the consistency and accuracy but also to compare the post dermatological aesthetic metrics.




\ifCLASSOPTIONcaptionsoff
\newpage
\fi

\bibliography{biblio.bib}
\bibliographystyle{IEEEtran}

\end{document}